\documentclass[11pt]{article}
\usepackage{emnlp2016}
\usepackage{times}
\usepackage{amsmath}
\usepackage{latexsym}
\usepackage{graphicx}
\usepackage{multirow}
\usepackage{float}
\usepackage{caption}

\usepackage{subfig}

\usepackage{hhline}
\usepackage{tcolorbox}

\emnlpfinalcopy
\def\emnlppaperid{585}

\title{A Neural Network for Coordination Boundary Prediction}

\author{Jessica Ficler \\
 Computer Science Department \\
 Bar-Ilan University \\
 Israel \\
 {\tt jessica.ficler@gmail.com} \\\And
 Yoav Goldberg \\
 Computer Science Department \\
 Bar-Ilan University \\
 Israel \\
 {\tt yoav.goldberg@gmail.com} \\}

\date{}

\begin{document}

\maketitle

\begin{abstract}
We propose a neural-network based model for coordination boundary prediction.
The network is designed to incorporate two signals: the similarity between
conjuncts and the observation that replacing the whole coordination phrase with
a conjunct tends to produce a coherent sentences. The modeling makes use of
several LSTM networks. The model is trained solely
on conjunction annotations in a Treebank, without using external resources.
We show improvements on predicting coordination boundaries on the PTB 
compared to two state-of-the-art parsers; as well as improvement over previous
coordination boundary prediction systems on the Genia corpus.
\end{abstract}

\section{Introduction}
Coordination is a common syntactic phenomena, appearing in 38.8\% of the sentences in the Penn Treebank (PTB) \cite{ptb}, and in 60.71\% of the sentences in the Genia Treebank \cite{ohta2002genia}.
However, predicting the correct conjuncts span remain one of the biggest
challenges for state-of-the-art syntactic parsers.
Both the Berkeley and Zpar phrase-structure parsers
\cite{petrov2006learning,zhang2011syntactic} achieve F1 scores of around 69\%
when evaluated on their ability to recover coordination boundaries on the PTB test set.
For example, in:

\vspace{5pt}
\noindent
\textit{``\small He has the government's blessing to [build churches] and [spread Unificationism] in that country."}
\noindent
\vspace{5pt}

\noindent the conjuncts are incorrectly predicted by both parsers:

\vspace{5pt}
\noindent
{\small Berkeley: \textit{``He has the government's blessing to [build churches] and [spread Unificationism in that country]."}}\\
{\small Zpar: \textit{``He [has the government's blessing to build churches] and [spread Unificationism in that country]."}}
\noindent
\vspace{5pt}

In this work we focus on coordination boundary prediction, and suggest a
specialized model for this task. We treat it as a ranking task, and learn
a scoring function
over conjuncts candidates such that the correct candidate pair
is scored above all other candidates.
The scoring model is a neural network with two LSTM-based components,
each modeling a different linguistic principle: (1) conjuncts tend to be
similar (``symmetry''); and (2) replacing the coordination phrase with each of the conjuncts
usually result in a coherent sentence (``replacement'').
The \emph{symmetry} component takes into account the conjuncts' syntactic structures,
allowing to capture similarities that occur in different levels of the syntactic
structure. 

The \emph{replacement} component considers the coherence of the sequence that is produced when connecting the participant parts.
Both of these signals are syntactic in nature, and are learned solely based on
information in the Penn Treebank. Our model substantially outperforms both the
Berkeley and Zpar parsers on the coordination prediction task, while using the
exact same training corpus.
Semantic signals (which are likely to be based on resources
external to the treebank) are also relevant for coordination disambiguation
\cite{kawahara2008coordination,hogan2007coordinate} and
provide complementary information. We plan to incorporate such signals in future
work.  

\section{Background}
Coordination is a very common syntactic construction in which several sentential elements
(called conjuncts) are linked. For example, in:

\vspace{5pt}
\noindent
\textit{``The Jon Bon Jovi Soul Foundation [was founded in 2006] and$_1$ [exists to combat issues that force (families) and$_2$ (individuals) into economic despair]."}
\noindent
\vspace{5pt}

\noindent The coordinator \textit{and$_1$} links the conjuncts surrounded with square brackets and the coordinator \textit{and$_2$} links the conjuncts surrounded with round brackets.

Coordination between NPs and between VPs are the most common, but
other grammatical functions can also be coordinated:
\textit{``[relatively active]$_{ADJP}$ but [unfocused]$_{ADJP}$"} ; \textit{``[in]$_{IN}$ and [out]$_{IN}$ the market"}.
While coordination mostly occurs between elements with the same syntactic
category, cross-category conjunctions are also possible:
(\textit{``Alice will visit Earth [tomorrow]$_{NP}$ or [in the next
decade]$_{PP}$''}). 
Less common coordinations involve non-constituent elements
\textit{``[equal to] or [higher than]''}, argument clusters (\textit{``Alice 
visited [4 planets] [in 2014] and [3 more] [since then]''}), and gapping
(\textit{``[Bob lives on Earth] and [Alice on Saturn]''}) ~\cite{dowty1988type}.

\subsection{Symmetry between conjuncts}
\label{sec:bcksym}
Coordinated conjuncts tend to be semantically related and have a similar syntactic structure. 
For example, in (a) and (b) the conjuncts include similar words (China/Asia, marks/yen) and have identical syntactic structures. 

\begin{figure}[H]
\centering
   \subfloat[]{\includegraphics[scale=0.47]{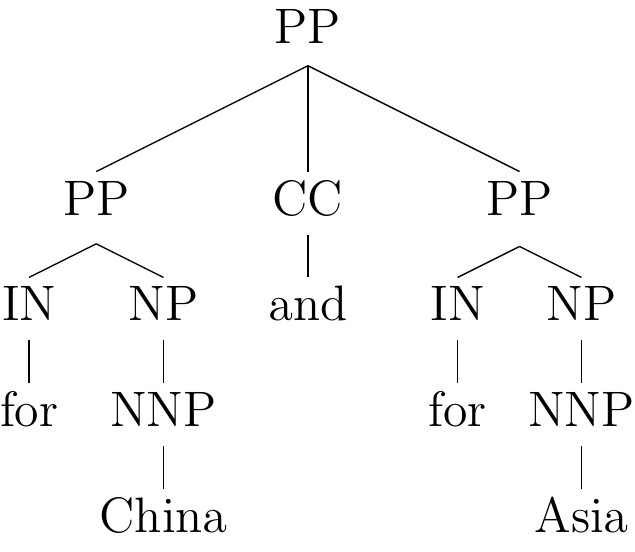} \label{fig:shared_ptb}}
   \hspace{0.4cm}
  \subfloat[]
  {\includegraphics[scale=0.47]{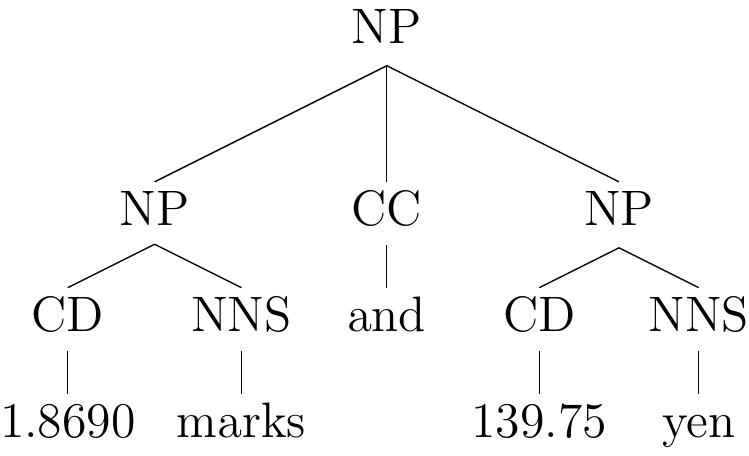}\label{fig:shared_our}}
\end{figure}
\eject
\noindent Symmetry holds also in larger conjuncts, such as in:

\begin{center}
(c) {\includegraphics[scale=0.3]{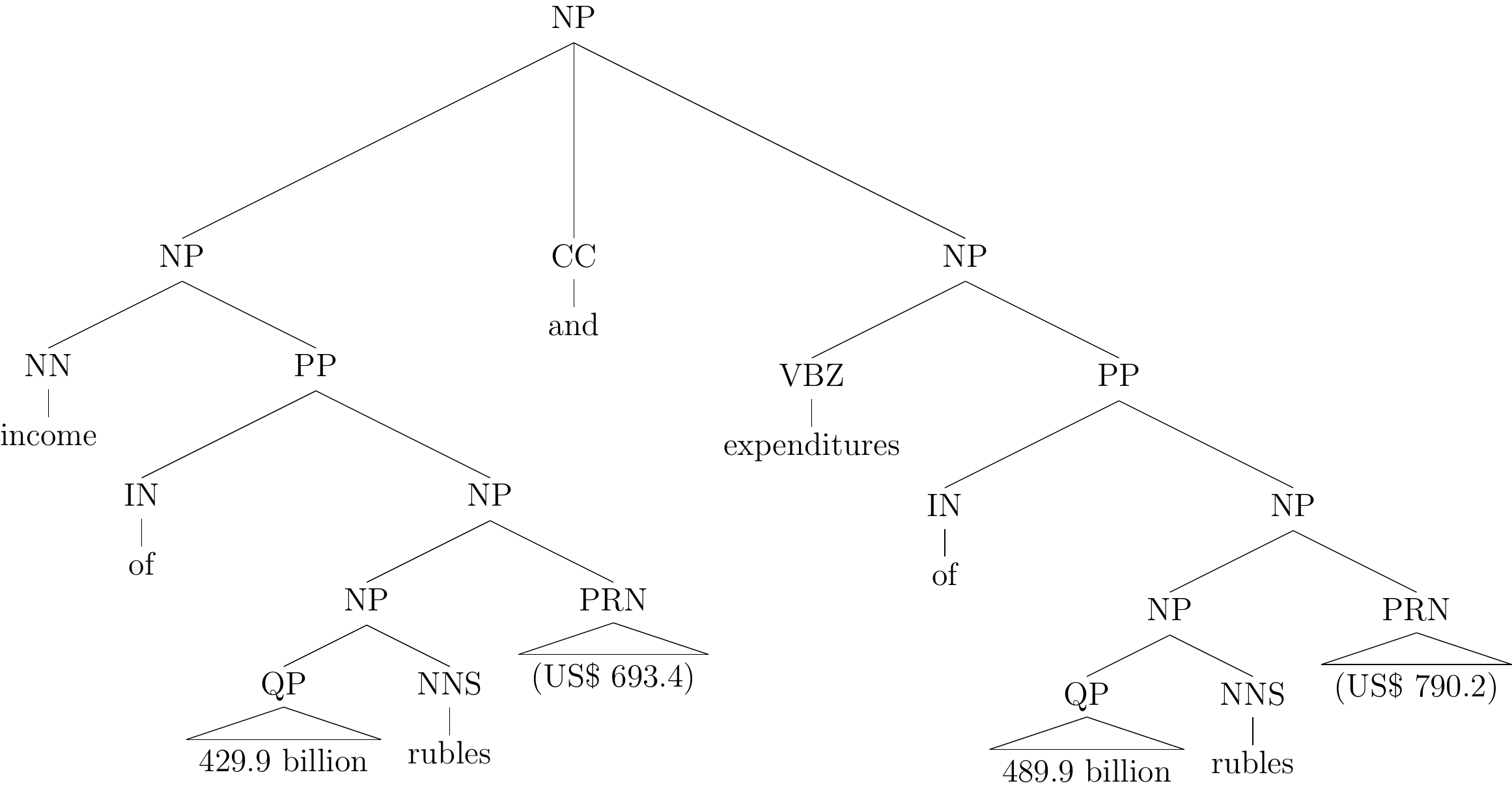}}
\end{center}

\noindent Similarity between conjuncts was used as a guiding principle in
previous work on coordination disambiguation \cite{hogan2007coordinate,shimbo2007discriminative,hara2009coordinate}.

\subsection{Replaceability}
\label{sec:bckrep}
Replacing a conjunct with the whole coordination phrase usually produce a coherent sentence \cite{huddleston2002cambridge}.
For example, in \textit{``Ethan has developed [new
products] and [a new strategy]''}, replacement results in: \textit{``Ethan
has developed new products"}; and \textit{``Ethan has developed a new
strategy"}, both valid sentences.
Conjuncts replacement holds also for conjuncts of different syntactic types,
e.g.:
\textit{``inactivation of tumor-suppressor genes, [alone] or
[in combination], appears crucial to the development of such scourges as
cancer.''}.

While both symmetry and replacebility are strong characteristics of coordination, neither
principle holds universally. Coordination between syntactically dissimilar
conjuncts is possible (``tomorrow and for the entirety of the next decade''),
and the replacement principle fails in cases of ellipsis, gapping and others (\textit{``The bank employs [8,000 people in Spain] and [2,000 abroad]''}). 

\subsection{Coordination in the PTB}
Coordination annotation in the Penn Treebank \cite{ptb} is inconsistent \cite{hogan2007coordinate} and lacks internal structure for NPs with nominal modifiers \cite{bies1995bracketing}.
In addition, conjuncts in the PTB are not explicitly marked. 
These deficiencies led previous works on coordination disambiguation \cite{shimbo2007discriminative,hara2009coordinate,hanamoto2012coordination} to use the Genia treebank of biomedical text \cite{ohta2002genia} which explicitly marks coordination phrases.
However, using the Genia corpus is not ideal since it is in a specialized domain and much smaller than the PTB.
In this work we rely on a version of the PTB released by Ficler and Goldberg \shortcite{ficler2016coordination} in which the above deficiencies are
manually resolved. In particular, coordinating elements, coordination phrases
and conjunct boundaries are explicitly marked with specialized function
labels.

\subsection{Neural Networks and Notation}
We use $w_{1:n}$ to indicate a list of vectors $w_1, w_2, \dots w_n$ and $w_{n:1}$
to indicate the reversed list.
We use $\circ$ for vector concatenation.
When a discrete symbol $w$ is used as a neural network's input,
the corresponding embedding vector is assumed.

A multi-layer perceptron (MLP) is a non linear
classifier. In this work we take MLP to mean a classifier with a single hidden
layer:
$MLP(x) = V\cdot g(Wx + b)$
where $x$ is the network's input, $g$ is an activation function such as
ReLU or Sigmoid, and $W$, $V$ and $b$ are trainable parameters.
Recurrent Neural Networks (RNNs) \cite{elman1990finding} allow the representation of arbitrary sized
sequences. 
In this work we use LSTMs \cite{hochreiter1997long}, a variant of RNN that was
proven effective in many NLP tasks.
$\text{LSTM}(w_{1:n})$ is the outcome vector resulting from feeding the sequence
$w_{1:n}$ into the LSTM in order. A bi-directional LSTM (biLSTM)
takes into account both the past
$w_{1:i}$ and the future $w_{i:n}$ when representing the element in position $i$:
\begin{equation}
\resizebox{0.9\hsize}{!}{$%
biLSTM(w_{1:n},i) = LSTM_F(w_{1:i})\circ LSTM_B(w_{n:i})$
} \nonumber
\end{equation}
where $LSTM_F$
reads the sequence in its regular order and $LSTM_B$ reads it in reverse.
\begin{figure}[t!]
\begin{scalebox}{0.8}
\centering
\begin{tcolorbox}
\begin{tabular}{l}
\bf Sentence: \\ 
\textit{\small And$_1$ the city decided to treat its guests more } \\
\textit{\small like royalty or$_2$ rock stars than factory owners.}  \\
\bf Expected output: \\
and$_1$:  NONE \\
or$_2$:   (11-11) {\small royalty} ; (12-13) {\small rock stars} \\
\hline
\bf Sentence: \\ 
\textit{\small The president is expected to visit Minnesota, New  } \\
\textit{\small York and$_1$ North Dakota by the end of the year.}  \\
\bf Expected output: \\
and$_1$:  (9-10) {\small New York} ; (12-13) {\small North Dakota}
\end{tabular}
\end{tcolorbox}
\end{scalebox}
\caption{\footnotesize{The coordination prediction task.}}
\label{fig:task}
\end{figure}

\begin{figure*}
{\includegraphics[scale=0.18]{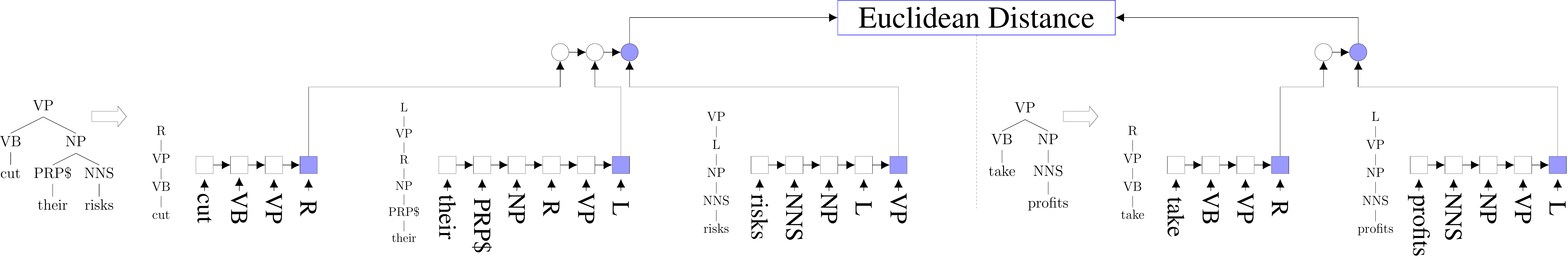}\label{}}
\caption{\footnotesize Illustration of the symmetry scoring component that
takes into account the conjuncts syntactic structures. Each conjunct tree is
decomposed into paths that are fed into the path-LSTMs (squares).
The resulting vectors are fed into the symmetry LSTM function (circles).
The outcome vectors (blue circles) are then fed into the euclidean distance
function.}
\label{fig:sym_net}
\end{figure*}

\section{Task Definition and Architecture}
\label{tsk:def}
Given a coordination word in a sentence, the coordination prediction
task aims to returns the two conjuncts that are connected by it, or \textsc{None}
if the word does not function as a coordinating conjunction of a relevant
type.\footnote{We consider \textit{and}, \textit{or}, \textit{but}, \textit{nor}
as coordination words. In case of more than two coordinated elements (conjuncts), we focus on the two
conjuncts which are closest to the coordinator.} Figure \ref{fig:task} provides
an example.

Our system works in three phases: first, we determine if the coordinating word
is indeed part of a conjunction of a desired type. We then extract a ranked list of
candidate conjuncts, where a candidate is a pair of spans of the form ($(i,j)$,
$(l,m)$).
The candidates are then scored and the highest scoring pair is returned.
Section \ref{sec:score} describes the scoring model, which is the main
contribution of this work. The coordination classification and candidate
extraction components are described in Section \ref{sec:supporting}.

\section{Candidate Conjunctions Scoring}
\label{sec:score}
Our scoring model takes into account two signals, symmetry between conjuncts and the possibility of replacing the whole coordination phrase with its participating conjuncts.

\subsection{The Symmetry Component}
As noted in Section \ref{sec:bcksym}, many conjuncts spans have similar syntactic
structure. 
However, while the similarity is clear to human readers, it is often not easy to formally define, such as in:
\begin{center}

\textit{``about/IN half/NN its/PRP\$ revenue/NN \\and/CC\\
more/JJR than/IN half/NN its/PRP\$ profit/NN"}
\end{center}

If we could score the amount of similarity between two spans, we could use that
to identify correct coordination structures. However, we do not know the
similarity function. We approach this by training the similarity function in a
data-dependent manner. Specifically, we train an encoder that encodes spans into
vectors such that vectors of similar spans will have a small Euclidean distance
between them. This architecture is similar to Siamese Networks, which are used for
learning similarity functions in vision tasks \cite{chopra2005learning}.

Given two spans of lengths $k$ and $m$ with corresponding vector sequences $u_{1:k}$ and $v_{1:m}$ we encode each sequences using an LSTM, and take the euclidean distance between the resulting representations:
\begin{equation}
\resizebox{0.9\hsize}{!}{$%
Sym(u_{1:k},v_{1:m}) = || LSTM(u_{1:k}) - LSTM(v_{1:m}) ||$
} \nonumber
\end{equation}
The network is trained such that the distance is minimized for compatible spans and large for incompatible ones in order to learn that vectors that represent correct conjuncts are closer than vectors that do not represent conjuncts.

What are the elements in the sequences to be compared?

One choice is to take the vectors $u_i$ to correspond to embeddings of the $i$th POS in the span.

This approach works reasonably well, but does not consider the conjuncts'
syntactic structure, which may be useful as symmetry often occurs on a higher
level than POS tags. For example, in:

\includegraphics[scale=0.43]{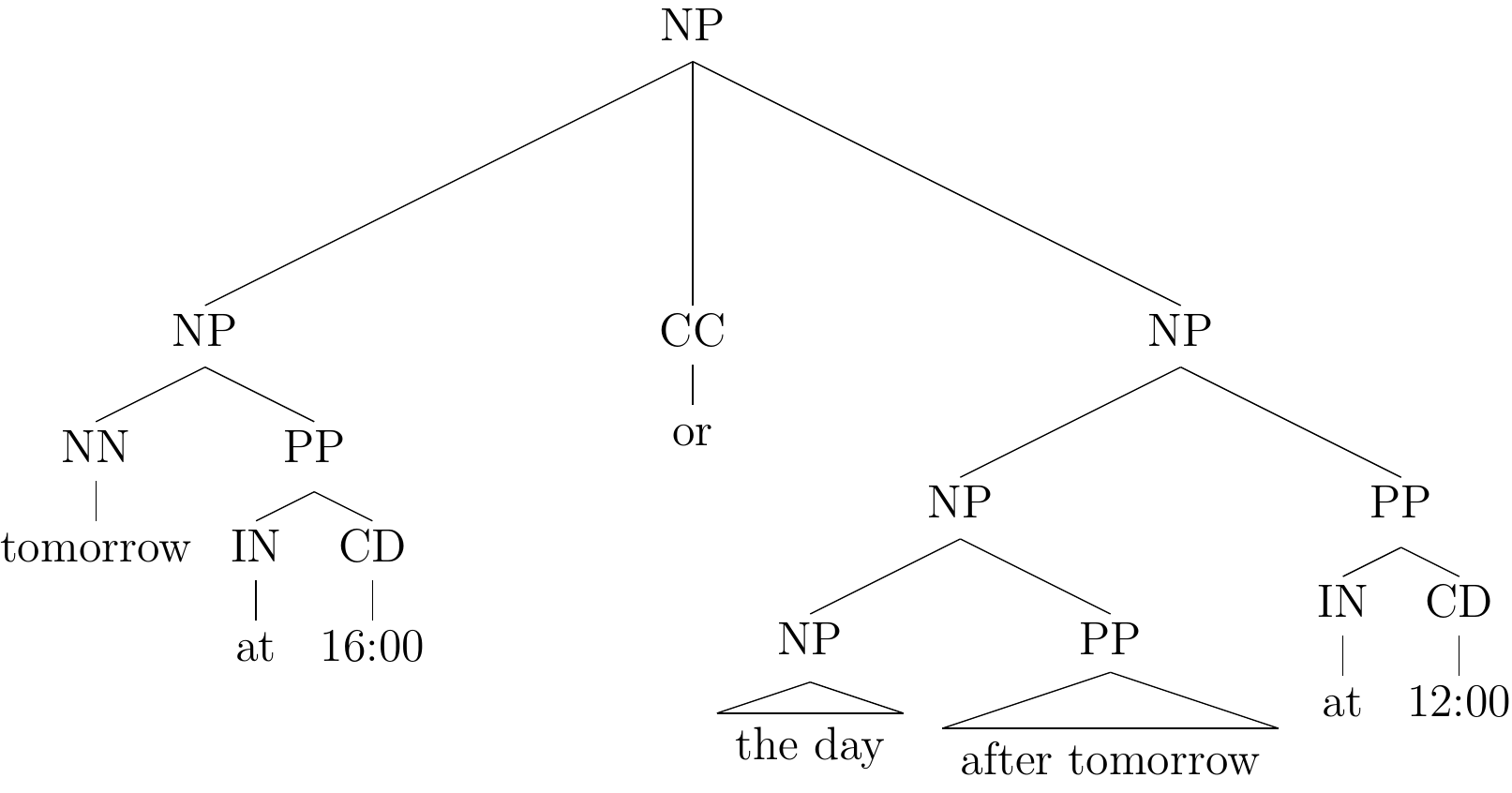}

\noindent the similarity is more substantial in the third level of the tree than in the POS level.

A way to allow the model access to higher levels of syntactic symmetry is to represent
each word as the projection of the grammatical functions from the word to the
root.\footnote{Similar in spirit to the spines used in Carreras et al. \shortcite{carreras2008tag} and Shen et al. \shortcite{shen2003using}.}

For example, the projections for the first conjunct in Figure \ref{fig:sym_net} are:
\begin{center}
\includegraphics[scale=0.5]{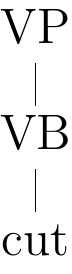} 
\hspace{1cm}
\includegraphics[scale=0.5]{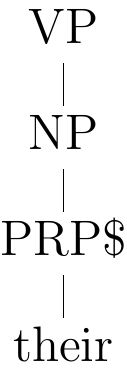} 
\hspace{1cm}
\includegraphics[scale=0.5]{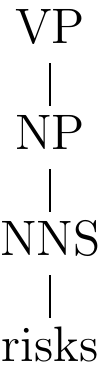} 
\end{center}

This decomposition captures the syntactic context of each word, but does not
uniquely determine the structure of the tree. To remedy this, we add to the
paths special symbols, $R$ and $L$, which marks the lowest common ancestors with the right and left words respectively.

These are added to the path above the corresponding nodes. For example consider the following paths which corresponds to the above syntactic structure:

\begin{center}
 \includegraphics[scale=0.5]{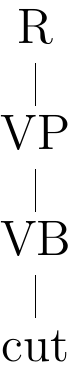} 
 \hspace{1cm}
 \includegraphics[scale=0.5]{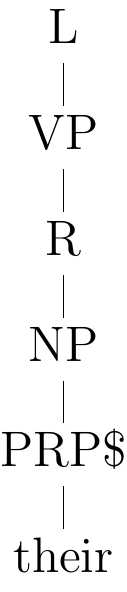} 
 \hspace{1cm}
 \includegraphics[scale=0.5]{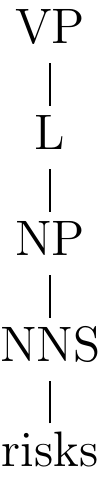}
\end{center}

\noindent The lowest common ancestor of \textit{``their"}  and \textit{``risks"} is NP. Thus, $R$ is added after NP in the path of \textit{``their"} and $L$ is added after NP in the path of \textit{``risks"}. 
Similarly, $L$ and $R$ are added after the VP in the \textit{``their"} and \textit{``cut"}
paths.

\begin{figure*}
\includegraphics[scale=0.7]{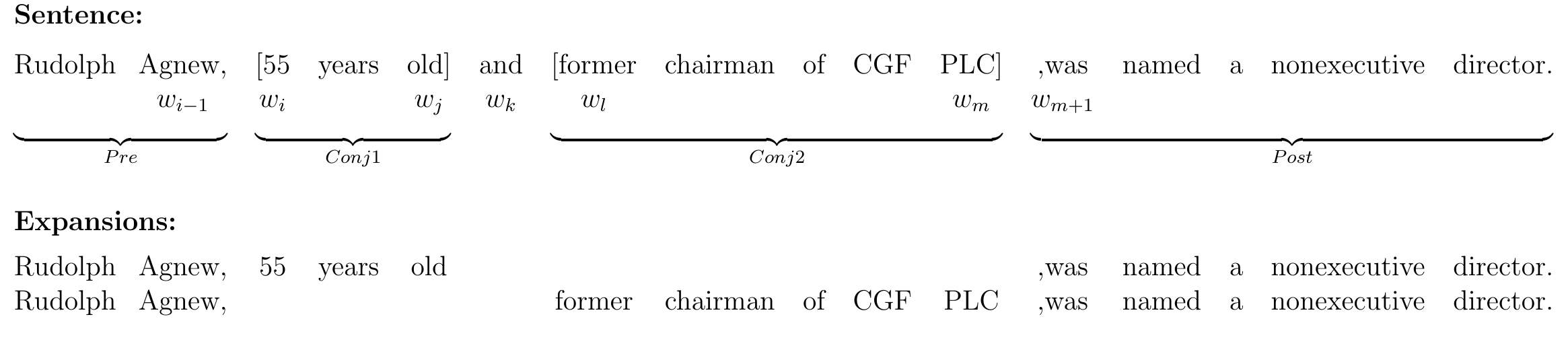}
\caption{\footnotesize The correct conjuncts spans of the coordinator \textit{and} in the sentence and the outcome expansions.}
\label{fig:rep}
\end{figure*}

The path for each word is encoded using an LSTM receiving vector embeddings of the elements in the path from the word to the root.
We then use the resulting encodings instead of the POS-tag embeddings as input to the LSTMs in the similarity function.

Figure \ref{fig:sym_net} depicts the complete process for the spans ``cut their
risks'' and ``take profits''.

Using syntactic projections requires the syntactic structures of the conjuncts.
This is obtained by running the Berkeley parser over the sentence and taking
the subtree with the
highest probability from the corresponding cell in the CKY chart.\footnote{The
parser's CKY chart did not include a tree for 10\% of the candidate spans, which have inside probability 0 and outside probability $>$ 0. For those, we obtained the syntactic structure by running the parser on the span words only.}
In both approaches,
the POS embeddings are initialized with vectors that are pre-trained by running
word2vec \cite{mikolov2013efficient} on the POS sequences in PTB training
set.

\subsection{The Replacement Component}
The replacement component is based on the observation that, in many cases, the
coordination phrase can be replaced with either one of its conjuncts while still
preserving a grammatical and semantically coherent sentence (Section
\ref{sec:bckrep})

When attempting such a replacement on incorrect conjuncts, the resulting
sentence is likely to be either syntactically or semantically
incorrect.
For example, in the following erroneous analysis:
\textit{``Rudolph Agnew, [55 years old] and [former chairman] of Consolidated Gold Fields PLC"}
replacing the conjunction with the first conjunct results in the
semantically incoherent sequence \textit{``Rudolph Agnew, 55 years old of
Consolidated Golden Fields, PLC''}.\footnote{While uncommon, incorrect conjuncts may also result in valid
sentences, e.g. \textit{``He paid \$ 7 for cold [drinks] and [pizza] that just came out
of the oven.''}}

Our goal is to distinguish replacements resulting from
correct conjuncts from those resulting from erroneous ones.
To this end, we focus on the \textit{connection
points}.
A connection point in a resulting sentence is the point where the sentence
splits into two sequences that were not connected in the original sentence.
For example, consider the sentence in Figure \ref{fig:rep}.
It has four parts, marked as \textit{Pre}, \textit{Conj1}, \textit{Conj2} and
\textit{Post}. Replacing the coordination phrase \textit{Conj1 and Conj2} with
\textit{Conj2} results in a connection point between \textit{Pre} and
\textit{Conj2}. Likewise, replacing the coordination phrase with \textit{Conj1}
results in connection point between \textit{Conj1} and \textit{Post}.

In order to model the validity of the connection points, we represent 
each connection point as the concatenation of a forward and reverse LSTMs
centered around that point. Specifically, for the spans in Figure \ref{fig:rep}
the two connection points are represented as:\\
$\scriptstyle LSTM_F(\texttt{Rudolph},...,\texttt{old})\circ
LSTM_B(\texttt{director},...,\texttt{was},\texttt{,})$\\ and\\
$\scriptstyle LSTM_F(\texttt{Rudolph},\texttt{Agnew},\texttt{,})\circ
LSTM_B(\texttt{director},...,\texttt{former})$

Formally, assuming words $w_{1:n}$ in a sentence with coordination at position
$k$ and conjuncts $w_{i:j}$ and $w_{l:m}$,\footnote{Usually $j=k-1$ and
$l=k+1$, but in some cases punctuation symbols may interfere.}

the connection points are between $w_{1:j}$ and $w_{m+1:n}$; and between $w_{1:i-1}$ and $w_{l:n}$.
The two connection points representations are then concatenated, resulting in a
\textit{replacement vector}:

\begin{equation}
\resizebox{0.97\hsize}{!}{$%
\begin{aligned}
\textsc{Repl}(w_{1:n}, i, j, l, m)= &\\
&\hspace{-2.5cm} \textsc{ConPoint}(w_{1:n}, i-1, l) \circ \textsc{ConPoint}(w_{1:n}, j, m+1)
\end{aligned}
$} \nonumber
\end{equation}
where: 
\begin{equation}
\resizebox{0.6\hsize}{!}{$%
\begin{aligned}
\textsc{ConPoint}(w_{1:n}, i ,j) = &\\
&\hspace{-2.5cm} LSTM_F(w_{1:i})\circ LSTM_B(w_{n:j})
\end{aligned}
$} \nonumber
\end{equation}

We use two variants of the replacement vectors, corresponding to two levels of
representation. The first variant is based on the sentence's words, while the
second is based on its POS-tags.

\subsection{Parser based Features}
In addition to the symmetry and replacement signals, we also incorporate some
scores that are derived from the Berkeley parser. As detailed in Section
\ref{sec:supporting}, a list of conjuncts candidates are extracted from the CKY
chart of the parser. The candidates are then sorted in descending order
according to the multiplication of inside and outside
scores of the candidate's spans:\footnote{Inside-Outside probabilities
\cite{goodman1998parsing} represent the probability of a span with a given
non-terminal symbol. The inside probability $I_{(N,i,j)}$ is the probability of generating words
$w_i,w_{i+1},...,w_{j}$ given that the root is the non-terminal $N$.
The outside probability $O_{(N,i,j)}$ is the probability of generating words $w_{1},w_{2},...,w_{i-1}$, the non-terminal $N$ and the words $w_{j+1},w_{j+2},...,w_{n}$ with the root $S$.} $I_{(i,j)}\times O_{(i,j)}\times I_{(l,m)}\times O_{(l,m)}$.
Each candidate $\{(i,j),(l,m)\}$ is assigned two numerical features based on this ranking: its position in
the ranking, and the ratio between its score and the score of the adjacent
higher-ranked candidate. We add an additional binary feature indicating whether the
candidate spans are in the 1-best tree predicted by the parser.
These three features are denoted as $Feats(i,j,l,m)$.

\subsection{Final Scoring and Training}
Finally, the score of a candidate $\{(i,j), (l,m)\}$ in a sentence with words
$w_{1:n}$ and POS tags $p_{1:n}$ is computed as:
\begin{equation}
\resizebox{0.6\hsize}{!}{$%
\begin{aligned}
    \textsc{Score}(w_{1:n},p_{1:n},\{(i,j),(l,m)\}) = &\\
&\hspace{-4cm}MLP(\\
&\hspace{-3.5cm} Sym(v_{i:j}^{Path} , v_{l:m}^{Path}) \\
&\hspace{-3.5cm} \circ Repl(w_{1:n},i,j,l,m) \\
&\hspace{-3.5cm} \circ Repl(p_{1:n},i,j,l,m) \\
&\hspace{-3.5cm} \circ Feats(i,j,l,m)\;\; )
\end{aligned}
$} \nonumber
\end{equation}
where $v_{i:j}^{Path}$ and $v_{l:m}^{Path}$ are the vectors resulting from the
path LSTMs, and $Sym$, $Repl$ and $Feats$ are the networks defined in Sections
4.1 -- 4.3 above.  The network is trained jointly, attempting to minimize a
pairwise ranking loss function, where the loss for each training case is given
by:
\begin{center}
$loss = max(0, 1 - (\hat{y} - y_g))$
\end{center}
where $\hat{y}$ is the highest scoring candidate and $y_g$ is the correct candidate.
The model is trained on all the coordination cases in Section 2--21 in the PTB.

\section{Candidates Extraction and Supporting Classifiers}
\label{sec:supporting}
\paragraph{Candidates Extraction}
We extract candidate spans based on the inside-outside probabilities assigned by
the Berkeley parser.
Specifically, to obtain candidates for conjunct span
we collect spans that are marked with COORD, are adjacent to the coordinating
word, and have non-zero inside or outside probabilities.  We then
take as candidates all possible pairs of collected spans.

On the PTB dev set, 
this method produces 6.25 candidates for each coordinating word on average and
includes the correct candidates for 94\% of the coordinations.

\paragraph{Coordination Classification}
We decide whether a coordination word $w_k$ in a sentence $w_{1:n}$ functions as
a coordinator by feeding the biLSTM vector centered around $w_k$ 
into a logistic classifier:
\begin{center}
$\sigma(v\cdot\text{biLSTM}(w_{1:n},k)+b)$.
\end{center}
The training examples are all the coordination words (marked with CC) in the PTB training set.
The model achieves 99.46 F1 on development set and 99.19 F1 on test set.

NP coordinations amount to about half of the coordination cases in the PTB,
and previous work is often evaluated specifically on NP coordination.
When evaluating on NP coordination, we depart from the unrealistic scenario used in most
previous work where the type of coordination is assumed to be known a-priori,
and train a specialized model for predicting the coordination type.
For a coordination candidate $\{(i,j),(l,m)\}$ with a coordinator $w_k$,
we predict if it is NP coordination or not by feeding a logistic classifier with
a biLSTM vector centered around the coordinator and constrained to the candidate
spans:
\begin{center}
$\sigma(v\cdot\text{biLSTM}(w_{i:m},k)+b)$.
\end{center}

The training examples are coordinations in the PTB training set, where
where a coordinator is considered of type NP if its head is labeled with NP or NX.
Evaluating on gold coordinations results in F1 scores of 95.06 (dev) and 93.89
(test).

\section{Experiments}
We evaluate our models on their ability to identify conjunction boundaries in
the extended Penn Treebank \cite{ficler2016coordination} and Genia Treebank
\cite{ohta2002genia}\footnote{http://www-tsujii.is.s.u-tokyo.ac.jp/GENIA}.

When evaluating on the PTB, we compare to the conjunction boundary
predictions of the generative Berkeley parser \cite{petrov2006learning} and the discriminative Zpar
parser \cite{zhang2011syntactic}.
When evaluating on the Genia treebank, we compare to the results of the
discriminative coordination-prediction model of Hara et al.
\shortcite{hara2009coordinate}.\footnote{Another relevant model in the
literature is \cite{hanamoto2012coordination}, however the results are not
directly comparable as they use a slightly different definition of conjuncts,
and evaluate on a subset of the Genia treebank, containing only trees that were
properly converted to an HPSG formalism.}

\begin{table}[t!]
\centering
\scalebox{0.76}{
\begin{tabular}{|l|c|c|c|c|c|c|}
\multirow{1}{*}{} & \multicolumn{3}{c|}{Dev} & \multicolumn{3}{c|}{Test} \\
\cline{2-7}
 & P & R &F& P & R &F\\
\cline{1-7}

 Berkeley  & 70.14& 70.72& 70.42
 & 68.52& 69.33& 68.92
\\
Zpar  & 72.21& 72.72& \bf 72.46
 &68.24& 69.42& 68.82
\\ 
Ours &72.34 &72.25 &   72.29
  &72.81&72.61  & \bf 72.7
\end{tabular}}

\caption{\footnotesize Coordination prediction on PTB (All coordinations).}
\label{tbl:ptb_all_coord}
\end{table}

\begin{table}[t!]
\centering
\scalebox{0.76}{
\begin{tabular}{|l|c|c|c|c|c|c|}
\multirow{1}{*}{} & \multicolumn{3}{c|}{Dev} & \multicolumn{3}{c|}{Test} \\
\cline{2-7}
 & P & R &F& P & R &F\\
\cline{1-7}

 Berkeley  &67.53 &70.93 & 69.18
 & 69.51& 72.61& 71.02
 \\
Zpar  & 69.14& 72.31& 70.68
 &69.81& 72.92& 71.33
 \\
Ours & 75.17& 74.82& \bf 74.99
  &76.91& 75.31 & \bf 76.1
\end{tabular}}
\caption{\footnotesize Coordination prediction on PTB (NP coordinations).}
\label{tbl:ptb_np_coord}
\end{table}

\subsection{Evaluation on PTB}

\textbf{Baseline} Our baseline is the performance of the Berkeley and Zpar parsers on the task presented in Section \ref{tsk:def}, namely: for a given coordinating word, determine the two spans that are being conjoined by it, and return \textsc{None} if the coordinator is not conjoining spans or conjoins spans that are not of the expected type.
We convert predicted trees to conjunction predictions by taking the two
phrases that are immediately adjacent to the coordinator on both sides (ignoring
phrases that contain solely punctuation).
For example, in the following Zpar-predicted parse tree the conjunct prediction
is (\textit{``Feb. 8, 1990"},\textit{``May 10, 1990"}).
\begin{center}
\includegraphics[scale=0.5]{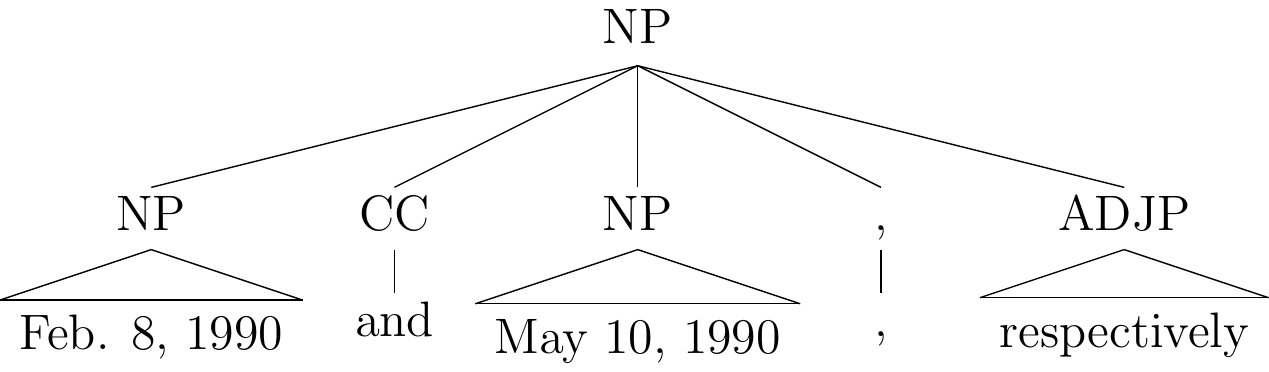}
\end{center}
Cases in which the coordination word is the left-most or right-most
non-punctuation element in its phrase (e.g. \texttt{(PRN (P -)(CC
and)(S it's been painful)(P -))}) are
considered as no-coordination (``None'').

We consider two setups. In the first 
we are interested in all occurrences of coordination, and in the second we focus on NP coordination. 
The second scenario requires typed coordinations. We take the type of a
parser-predicted coordination to be the type of the phrase immediately
dominating the coordination word.

\noindent\textbf{Evaluation Metrics}
We measure precision and recall compared to the gold-annotated coordination
spans in the extended PTB, where an example is considered correct if both conjunct
boundaries match exactly.  When focusing on NPs coordinations, the type of the phrase above the CC level
must match as well, and phrases of type NP/NX are considered as NP coordination.

\noindent\textbf{Results} Tables (\ref{tbl:ptb_all_coord}) and
(\ref{tbl:ptb_np_coord}) summarize the results. The Berkeley and Zpar parsers
perform similarly on the coordination prediction task. Our proposed model
outperforms both parsers, with a test-set $F_1$ score of 72.7 (3.78 $F_1$ points
gain over the better parser) when considering all coordinations, and test-set
$F_1$ score of 76.1 (4.77 $F_1$ points gain) when considering NP coordination.

\begin{figure*}
\centering
   \includegraphics[scale=0.61]{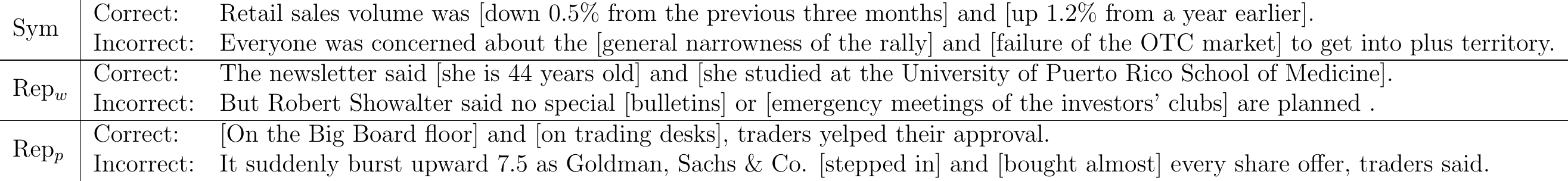} \label{examp1}

\caption{\footnotesize Correct in incorrect predictions by the individual
components.}
\label{fig:examp}
\end{figure*}

\subsection{Evaluation on Genia}
\label{sec:genia}
To compare our model to previous work, we evaluate also on the Genia treebank
(Beta), a collection of constituency trees for 4529 sentences from Medline
abstracts.
The Genia treebank coordination annotation explicitly marks coordination
phrases with a special function label (COOD),
making the corpus an appealing resource for previous work on coordination boundary
prediction
\cite{shimbo2007discriminative,hara2009coordinate,hanamoto2012coordination}.

Following Hara et al. \shortcite{hara2009coordinate}, we evaluate the models'
ability to predict the span of the entire coordination phrase, disregarding the
individual conjuncts. For example, in 
\textit{``My plan is to visit Seychelles, ko Samui and Sardinia by the end of
the year"} the goal is to recover \textit{``Seychelles, ko Samui and Sardinia"}.
This is a recall measure.
We follow the exact protocol of Hara et al. \shortcite{hara2009coordinate} and
train and evaluate the model on 3598 coordination phrases in Genia Treebank Beta
and report the micro-averaged results of a five-fold cross validation run.\footnote{We
thank Kazuo Hara for providing us with the exact details of their splits.}
As shown by Hara et al. \shortcite{hara2009coordinate}, syntactic parsers do not
perform well on the Genia treebank. Thus, in our symmetry component we opted to
not rely on predicted tree structures, and instead use the simpler option of
representing each conjunct by its sequence of POS tags.
To handle coordination phrases with more than two conjuncts, we extract
candidates which includes up to 7 spans and integrate the first and the last
span in the model features. Like Hara et al., we use gold POS.\\
\textbf{Results} Table
\ref{tbl:g2} summarizes the results.
Our proposed model achieves Recall score of 64.14 (2.64 Recall points gain over Hara et al.) and significantly improves the score of several coordination types.

\begin{table}[t]
\centering
\scalebox{0.76}{
\begin{tabular}{|c|c|c|c|}
COOD& \# &Our Model & Hara et al. \\ \hline
Overall & 3598 & \bf 64.14 & 61.5 \\ \hline
NP & 2317 & \bf 65.08& 64.2\\
VP & 465 & \bf 71.82 & 54.2\\
ADJP & 321 & 74.76 & \bf 80.4\\
S & 188 & 17.02 & \bf 22.9\\
PP & 167 & 56.28 & \bf 59.9\\
UCP & 60 & \bf 51.66 & 36.7\\
SBAR & 56 & \bf 91.07& 51.8\\
ADVP & 21 & 80.95 & \bf 85.7\\
Others & 3 & 33.33 & \bf  66.7\\
\end{tabular}}
\caption{\footnotesize Recall on the Beta version of Genia corpus. Numbers for Hara et al. are
taken from their paper.}
\label{tbl:g2}
\end{table}

\subsection{Technical Details}
The neural networks (candidate scoring model and supporting classifiers) are
implemented using the pyCNN
package.\footnote{https://github.com/clab/cnn/tree/master/pycnn}.

In the supporting models we use words embedding of size 50 and the Sigmoid activation function. 
The LSTMs have a dimension of 50 as well.
The models are trained using SGD for 10 iterations over the train-set, where 
samples are randomly shuffled before each iteration. We choose the model with
the highest F1 score on the development set.

All the LSTMs in the candidate scoring model have a dimension of 50. The input vectors for the symmetry LSTM is of size 50 as well.
The MLP in the candidate scoring model uses the Relu activation function, and
the model is trained using the Adam optimizer.
The words and POS embeddings are shared between the symmetry and replacment components. The syntactic label embeddings are for the path-encoding LSTM,
We perform grid search with
5 different seeds and the following: [1] MLP hidden layer size (100, 200,
400); [2] input embeddings size for words, POS and syntactic labels (100, 300).
We train for 20 iterations over the train
set, randomly shuffling the examples before each iteration.
We choose the model that achieves the highest F1 score on the dev set.

\section{Analysis}

\begin{table}[t!]
\centering
\scalebox{0.76}{
\begin{tabular}{|l||c|c|c||c|c|c|}
 & \multicolumn{3}{c||}{All types} & \multicolumn{3}{c|}{NPs} \\
\cline{2-7}
  & P & R &F& P & R &F\\
\cline{1-7}
  Sym & 67.13	&67.06&	67.09 & 69.69&	72.08&	70.86
 \\
        Rep$_p$ &69.26	&69.18&	69.21&
 69.73&	71.16&	70.43
 \\
         Rep$_w$ & 56.97	&56.9	&56.93&
 59.78&	64.3&	61.95
\\ 
  Feats & 70.92&	70.83&	70.87&
 72.23&	73.22	&72.72
\\ 

  Joint &72.34 &72.25 & \bf  72.29
  & 75.17& 74.82&  \bf 74.99\\
\end{tabular}}
\caption{\footnotesize Performance of the individual components on PTB section
22 (dev). Sym: Symmetry. Rep$_p$: POS replacement. Rep$_w$: Word replacement. Feats: features extracted from Berkeley parser.
Joint: the complete model.}
\label{tbl:ptb_partial}
\end{table}

Our model combines four signals: symmetry, word-level replacement, POS-level replacement and features from Berkeley parser. Table \ref{tbl:ptb_partial}
shows the PTB dev-set performance of each sub-model in isolation. On their own, each of the
components' signals is relatively weak, seldom outperforming the parsers. However, they
provide complementary information, as evident by the strong performance of the joint
model. Figure \ref{fig:examp} lists correct and incorrect predictions by each of
the components, indicating that the individual models are indeed capturing the patterns
they were designed to capture -- though these patterns do not always lead to
correct predictions.

\section{Related Work}

The similarity property between conjuncts was explored in several previous works on
coordination disambiguation.
Hogan \shortcite{hogan2007coordinate} incorporated this principle in a
generative parsing model by changing the generative process of coordinated NPs to condition on
properties of the first conjunct when generating the second one. 

Shimbo and Hara \shortcite{shimbo2007discriminative} proposed a discriminative
sequence alignment model to detect similar conjuncts. They focused on disambiguation of
non-nested coordination based on the learned edit distance between two
conjuncts. 
Their work was extended by Hara et al. \shortcite{hara2009coordinate} to handle nested coordinations as well.
The discriminative edit distance model in these works is similar in spirit to
our symmetry component, but is restricted to sequences of POS-tags, and makes
use of a sequence alignment algorithm. We compare
our results to Hara et al.'s in Section \ref{sec:genia}.
Hanamoto et al. \shortcite{hanamoto2012coordination} extended the previous method
with dual decomposition and HPSG parsing.
In contrast to these symmetry-directed efforts, 
Kawahara et al. \shortcite{kawahara2008coordination} focuses on the dependency
relations that surround the conjuncts. This kind of semantic information
provides an additional signal which is complementary to the syntactic signals
explored in our work. Our neural-network based model easily supports
incorporation of additional signals, and we plan to explore such semantic
signals in future work.

\section{Conclusions}
We presented an neural-network based model for resolving conjuncts boundaries.
Our model is based on the observation that (a) conjuncts tend to be similar and
(b) that replacing the coordination phrase with a conjunct results in a coherent sentence.
Our models rely on syntactic information and do not incorporate resources
external to the training treebanks, yet improve over state-of-the-art parsers on
the coordination boundary prediction task.

\section*{Acknowledgments}
This work was supported by The Israeli Science Foundation (grant number 1555/15)
as well as the German Research Foundation via the
German-Israeli Project Cooperation (DIP, grant DA 1600/1-1).

\bibliography{bib}
\bibliographystyle{emnlp2016}

\end{document}